\documentclass[conference]{IEEEtran}
\IEEEoverridecommandlockouts
\usepackage{cite}
\usepackage{amsmath,amssymb,amsfonts}
\usepackage{graphicx}
\usepackage{textcomp}
\usepackage{xcolor}
\usepackage{algorithmic}

\usepackage{array}
\usepackage{booktabs}
\usepackage{multirow}
\usepackage{diagbox}
\usepackage{subfig} 

\usepackage{stfloats} 
\allowdisplaybreaks

\usepackage{soul}
\sethlcolor{green}
\usepackage{amsthm}

\begin{document}
\title{\LARGE \bf
 Empirical Prediction of Pedestrian Comfort in Mobile Robot–Pedestrian Encounters
 \thanks{\textcolor{red}{Accepted to IEEE ICRA 2026. This is the author-prepared version of the accepted paper and will be removed after the conference.}}
}
\title{Empirical Prediction of Pedestrian Comfort in Mobile Robot–Pedestrian Encounters
{\footnotesize \textsuperscript{}}
\thanks{This work was supported in part by the National Science and Technology Council (NSTC), Taiwan, under Grant NSTC 114-2628-E-006-010 and NSTC 114-2218-E-006-021.}
\thanks{National Cheng Kung University's Institutional Review Board (IRB) reviewed and approved all the study procedures.}
\thanks{The authors are with the Department of Mechanical Engineering, National Cheng Kung University, Tainan, Taiwan. (e-mail: \texttt{yliu@mail.ncku.edu.tw})}
\thanks{\textcolor{red}{Accepted to IEEE ICRA 2026. This is the author-prepared version of the accepted paper and will be removed after the conference.}}
}


\author{\IEEEauthorblockN{Alireza Jafari, Hong-Son Nguyen, and Yen-Chen Liu$^*$}}

\markboth{2026 IEEE International Conference on Robotics and Automation (ICRA)  
June 1-5, Vienna, Austria}%
{Shell \MakeLowercase{\textit{et al.}}: Bare Demo of IEEEtran.cls for IEEE Journals}

\maketitle

\begin{abstract}
Mobile robots joining public spaces like sidewalks must care for pedestrian comfort.
Many studies consider pedestrians' objective safety, for example, by developing collision avoidance algorithms, but not enough studies take the pedestrian's subjective safety or comfort into consideration. 
Quantifying comfort is a major challenge that hinders mobile robots from understanding and responding to human emotions.
We empirically look into the relationship between the mobile robot-pedestrian interaction kinematics and subjective comfort.
We perform one-on-one experimental trials, each involving a mobile robot and a volunteer.
Statistical analysis of pedestrians' reported comfort versus the kinematic variables shows moderate but significant correlations for most variables.
\textcolor{black}{Based on these empirical findings, we design three comfort} estimators/predictors derived from the minimum distance, the minimum projected time-to-collision, and a composite estimator.
The composite estimator employs all studied kinematic variables and reaches the highest prediction rate and classifying performance among the predictors.
The composite predictor has an odds ratio of 3.67.
In simple terms, when it identifies a pedestrian as comfortable, it is almost 4 times more likely that the pedestrian is comfortable rather than uncomfortable.
The study provides a comfort quantifier for incorporating pedestrian feelings into path planners for more socially compliant robots.
\end{abstract}


\section{Introduction}
Social robots are entering human-populated areas such as sidewalks, hospitals, and airports, and cause discomfort, anxiety, and other moral and ethical concerns~\cite{Furlanis2024}.
Most path-planning algorithms, for example,~\cite{Perez2021}, prioritize objective safety over subjective safety or comfort, perhaps due to the complexity of quantifying human emotions for the controller.
Without quantifying pedestrian anxiety, a mobile robot cannot understand pedestrians' feelings, predict their responses, and plan a socially acceptable path~\cite{Kapoor2023}.
Neglecting nuanced human factors yields socially awkward behavior patterns and walkers' discomfort~\cite{Shiomi2014}.
\par
Mobile robots must respect pedestrian comfort in public spaces, specifically on sidewalks~\cite{Jafari2024-3-soro}. 
Multiple factors affect the human-robot interaction's pleasantness, such as the robot's shape, size, and noise~\cite{kanda2017,bartneck2024} and the robot's personality~\cite{Moujahid2023}.
Another example is the interaction kinematics.
Safe and pleasant pedestrian walks require the robot to understand how the movement patterns affect the pedestrians' feelings~\cite{agrawal2025}.
Among the kinematic variables, previous research primarily uses the minimum distance between the robot and the human, and only at low speeds.
Other kinematic variables are vastly unexplored, e.g., speed, lateral distance, time-to-collision, and trajectory curvature.
\par
A novelty of \textcolor{black}{this} paper is the empirical study of the neglected kinematic variables' effects on pedestrian comfort.
During multiple one-on-one hallway trials, we record the trajectories of a mobile robot and a pedestrian's reported comfort.
We statistically analyze the correlations of six kinematic variables with the pedestrian reports.
The kinematic variables are robot speed, minimum distance, lateral distance, maximum curvature, minimum PTTC, and distance at minimum PTTC.
The results show that the minimum PTTC has the strongest correlation, followed by minimum distance and speed; the difference between the last two is not definitive for ranking.
\par
\textcolor{black}{Another contribution of this research} is the design of empirical comfort predictors. 
Using the recorded data and manual binning, we construct three predictors for comfort estimations using the kinematic variables.
The minimum distance predictor is selected because it is the most popular in research, possibly due to the simplicity of the measurements in practice.
The PTTC-based estimator is chosen because of its strongest correlation among the variables.
The composite estimator combines the results from all kinematic variables, aiming to capture as many comfort aspects of the pedestrian as possible.
The composite estimator has the highest $F_1$ score, followed by the minimum distance and PTTC-based predictors, respectively.
\par
We would like to highlight the point that the composite estimator has an odds ratio of 3.67.
In other words, when the estimator detects a comfortable pedestrian, it is nearly four times more likely that the pedestrian is actually comfortable.
The comfort evaluation framework developed in this work has broad implications for path-planning algorithms.
It provides a way for the algorithms to estimate pedestrian comfort, allowing them to incorporate the walkers' feelings into their cost functions.
Therefore, the estimator enables path planners to detect comfortable paths from the generated trajectories, for example, in reinforcement learning studies like~\cite{Lu2025, Kapoor2023}.
\par
The paper structure is as follows.
\textcolor{black}{Section~\ref{sec:lit} reviews the literature and positions the contribution.}
Section~\ref{sec:kin} introduces the kinematic variables.
Section~\ref{sec:exp} details the design of the experiments.
Section~\ref{sec:res} presents the results, discusses the statistics, designs the predictors, and details the limitations.
Section~\ref{sec:con} outlines the paper and suggests follow-ups.
\section{Related works}\label{sec:lit}
Human-robot spatial interaction originated from Hall's research on proxemics~\cite{Hall1963}.
His classification of human personal space zones has been extensively adopted in human-robot interaction research.
For example, Xu et al. study proxemics application to quadruped robot-human interaction~\cite{Xu2025}.
They conclude that moving sideways or gazing at the pedestrian encourages the walker to maintain further distance from the robot.
Moreover, Shiomi et al. apply the Social Force Model (SFM), a distance-based pedestrian model, to a mobile in a shopping mall~\cite{Shiomi2014}.
The pedestrians perceive the mobile robot as more comfortable than a traditional time-varying dynamic window method.
Augmenting SFM, Ægidius et al. add behavioral patterns into SFM and improve quadruped interactions with pedestrians~\cite{AEgidius2024}.
\par
Robot speed also affects human comfort.
Neggers et al. show that the higher the robot speed, the more passing distance pedestrians prefer and the less comfort they report~\cite{Neggers2022-Fron, Neggers2022-SORO}.
In addition, MacArthur et al. performed experiments using a mobile robot with multiple volunteers self-reporting their trust~\cite{MacArthur2017}. 
Their statistically significant analysis shows that the participants report lower trust in faster robots. 
Song et al. investigate the speed of the robot and the signaling distance effects on the perceived comfort for delivery robots~\cite{Song2025}.
The participants watch pre-recorded videos and answer questions afterward.
The volunteers feel less comfortable as the speed increases.
\par
Lateral distance and curvature are other factors influencing walkers' comfort.
Pedestrians maintain certain distances from the robot when they pass by. 
Neggers et al. report increasing discomfort and intensified avoidance behavior when the lateral distance decreases~\cite{Neggers2022-Fron}.
In addition, the robot's trajectory curvature, as a metric of movement smoothness, influences pedestrian anxiety. 
Greenberg et al. study the sharpness of the curvature on human emotional states, such as pleasure and arousal, by a virtual reality environment~\cite{Greenberg2025}.
The participants prefer moderate curvatures.
\par
Besides spatial geometry, pedestrian comfort depends on temporal variables, too.
In e-scooter-pedestrian interactions, Projected Time-to-Collision (PTTC) correlates with walkers' reported comfort~\cite{jafari2023-ifac}.
Jafari and Liu perform e-scooter-pedestrian trials in a hallway setup and report especially stronger correlations at very low PTTC corresponding to high relative speeds~\cite{Jafari2024-2-natcom}.
Mobile robots are significantly slower than e-scooters.
Thus, the correlations may be weaker; nonetheless, we expect the effect to exist to some extent in mobile robot-human interactions, as well.
\par
Overall, many studies still use Hall's research for interpersonal spaces developed for human-human encounters and apply it to human-robot interactions.
Although the minimum distance to the robot plays a role in human discomfort, it is not the only factor. 
In addition, the zones defined for human-human interactions differ from human-robot's.
For example, the intimate zone of 0.45 m is different when the interacting agent is a robot~\cite{Neef2023}.
Recently, researchers noticed the gap and modified the zones~\cite{Neggers2022-SORO, Neggers2022-Fron} or introduced new terms into the comfort estimations~\cite{Greenberg2025, Greenberg2025-2}.
In addition, other kinematic factors, spatial and temporal, influence the comfort that are rarely investigated for mobile robot-human interactions.
\par
\section{Kinematic Variables}\label{sec:kin}
This paper studies the relationship between pedestrians' subjective comfort, denoted as $S$, and objective kinematic measures during robot-pedestrian interactions in hallway environments.
Understanding pedestrian comfort is the primary challenge in social robotics. 
Currently, most researchers use the distance to the pedestrian and the degree of personal space violation in comfort quantification, ignoring other key variables.
We empirically study multiple kinematic variables' influence on walkers' comfort.
\par
The kinematic variables are the robot's speed, the minimum distance, the lateral distance, the maximum trajectory curvature, the minimum PTTC, and the distance at the moment of minimum PTTC. 
The variables are selected from prior research in human-robot interaction and proxemics theory.
Each variable captures a distinct aspect of the robot-pedestrian interaction that has been theoretically linked to comfort perception: spatial relationships (minimum distance, lateral distance), temporal dynamics (PTTC), and behavioral adaptations (trajectory curvature, robot speed).
\par
\textbf{Robot speed} is a key variable affecting pedestrian comfort. 
Previous research studies its effect on pedestrian comfort at low speeds ($\leq1~m/s$), while in this paper, we test significantly higher speeds.
At lower speeds, the randomness in human danger interpretations reflected in questionnaires obscures possible correlations.
Therefore, we select significantly higher speeds to push the pedestrians to their limits, where the responses are more centralized.
During the experiments, we arbitrarily assign one of the following two speed conditions to each participant:
\begin{itemize}
    \item \textbf{R14:} The robot moves at a constant speed of 1.4 m/s throughout the trial.
    \item \textbf{R28:} The robot moves at a constant speed of 2.8 m/s throughout the trial.
\end{itemize}
The robot's speed is kept constant; however, it steers away from the pedestrian during the avoidance maneuver.
\par
\begin{figure}[t]
\centering
\subfloat[]{\includegraphics[width=0.24\textwidth]{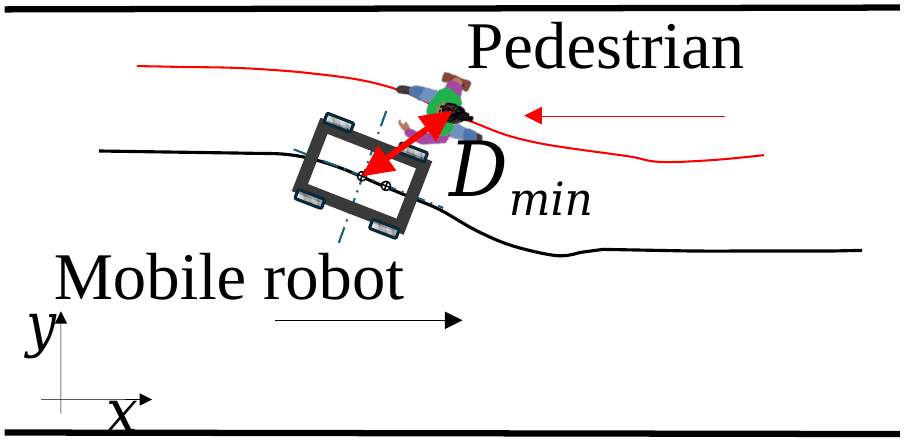}}
\subfloat[]{\includegraphics[width=0.24\textwidth]{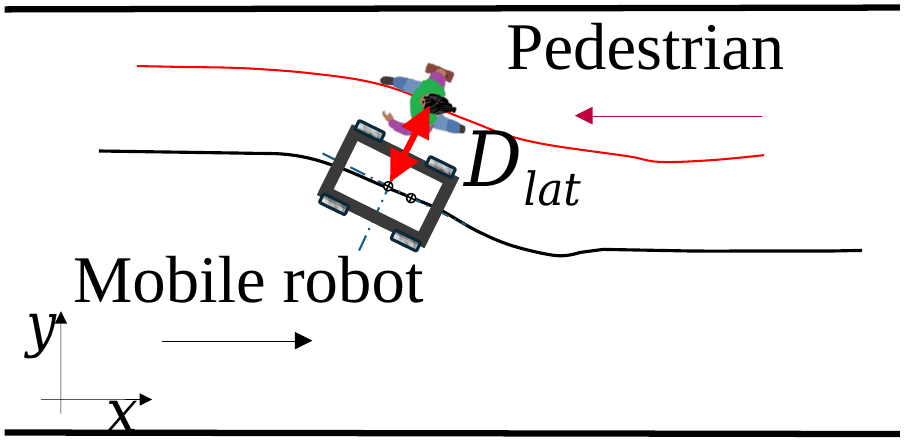}}
\\[2mm] 
\subfloat[]{\includegraphics[width=0.24\textwidth]{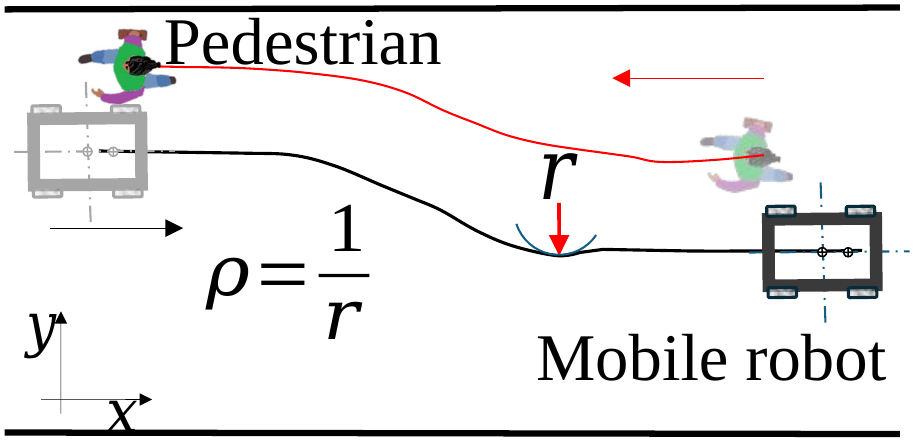}}
\subfloat[]{\includegraphics[width=0.24\textwidth]{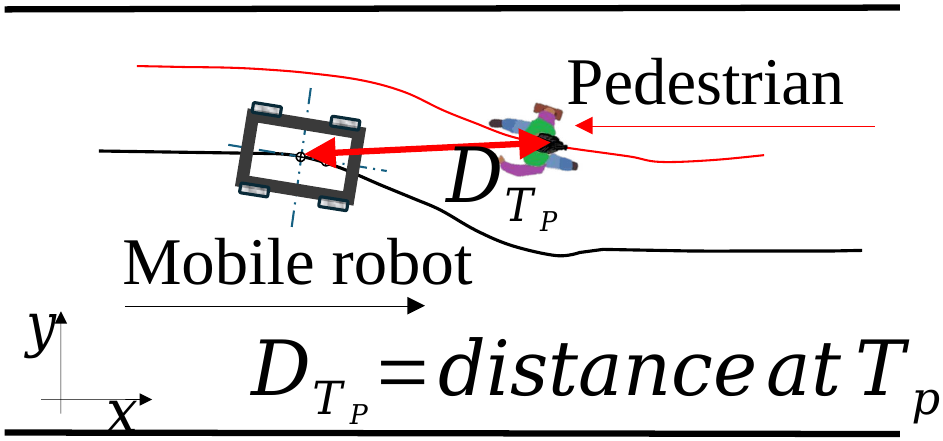}}
\caption{Spatial kinematic variables definitions.
(a) The minimum of the relative Euclidean distance between the robot and the pedestrian during a trial, $D_{min}$.
(b) The perpendicular distance between the robot and the pedestrian at the passing moment, $D_{lat}$. 
(c) The maximum curvature of the robot's trajectory during a trial, $\rho$.
(d) the distance between the robot and the pedestrian at the moment of $T_P$, $D_{T_P}$.}
\label{fig:variable_intro}
\end{figure}

\textbf{Minimum distance} is the primary predictor of pedestrian comfort that robotics researchers adopted from proxemics theory.
In this research, we evaluate its correlation with reported comfort alongside other kinematic variables.
~During the trials, we collect the relative Euclidean distance between the robot and the pedestrian.
Its minimum, $D_{min}$, is the most well-known variable in similar human comfort studies; see Fig.~\ref{fig:variable_intro}(a).
\par
\textbf{Lateral distance} is how close the robot comes to the pedestrian sideways as opposed to frontal confrontation.
It is also closely related to proxemic theory.
In all trials, the robot passes the human. 
The perpendicular distance between the robot and the pedestrian at the passing moment is the lateral distance $D_{lat}$; see Fig.~\ref{fig:variable_intro}(b).
\par
\textbf{Maximum curvature} is the maximum curvature along the robot’s trajectory during the interaction with a pedestrian.
Recent research highlights the curvature role in comfort estimation and suggests an optimal curvature~\cite{Greenberg2025}.
We report similar trends at higher speeds.
Fig.~\ref{fig:variable_intro}(c) shows the trajectory of the robot during a trial, where $r$ is its minimum radius and $\rho=\frac{1}{r}$ denotes the maximum curvature.
\par
\textbf{Minimum PTTC} is a subjective safety indicator in e-scooter-pedestrian interactions~\cite{Jafari2024-2-natcom}.
The research correlates the reported discomfort with the minimum PTTC.
We calculate PTTC evolution during each trial and use its minimum $T_p$ as a key temporal variable in mobile robot-pedestrian interactions.
Since mobile robots move more slowly than e-scooters (around 1 m/s compared to 2-5 m/s), the randomness in human danger interpretation may dilute the results.
Thus, the extension to mobile robots weakens the correlation; however, we are interested in whether it remains significant.
\par
\textbf{Distance at minimum PTTC} is a new variable introduced in this paper.
~Following the previous variable, the moment of minimum PTTC $T_P$ is a critical instant in pedestrian comfort assessment.
Fig.~\ref{fig:variable_intro}(d) shows the distance between the robot and the pedestrian at the moment, $D_{T_P}$.
We examine whether the variable affects pedestrian comfort and plays a complementary role alongside the other variables.
\par
We simultaneously measure and calculate these variables during each trial and statistically study their correlation with the reported subjective comfort.
The variables are interrelated because of their interconnected nature.
Nevertheless, each variable captures distinct aspects of the encounter. 
Subsequent statistical analysis examines both individual variable effects and their combined predictive power for comfort assessment.
Using the variables' empirical relation to reported comfort, we construct pedestrian comfort predictors and statistically compare their performance.
\par
\section{Experiments}\label{sec:exp}
This section focuses on the experimental methodology designed for the systematic collection of these kinematic data while recording participant comfort responses.
It includes the setup details, such as the hardware, and the design of the experiments, including the scenarios and the participant selection.
All the experimental procedures follow ethical guidelines and are approved by the Institutional Review Board (IRB) at the research institutions.
\par
\subsection{Experiment setup}\label{sec:exp:setup}
\begin{figure}[t]
    \centering
    \subfloat[]{\includegraphics[width=0.215\textwidth]{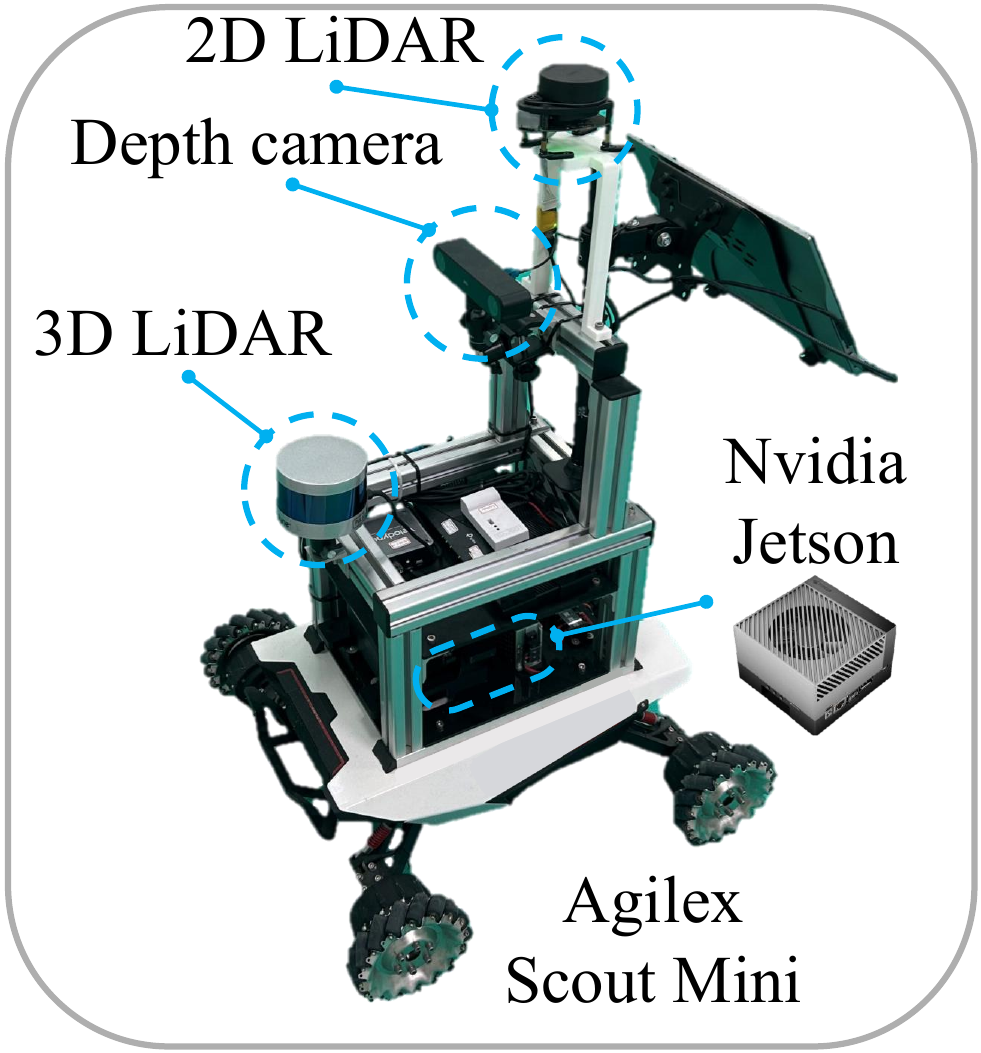}}
    \subfloat[]{\includegraphics[width=0.265\textwidth]{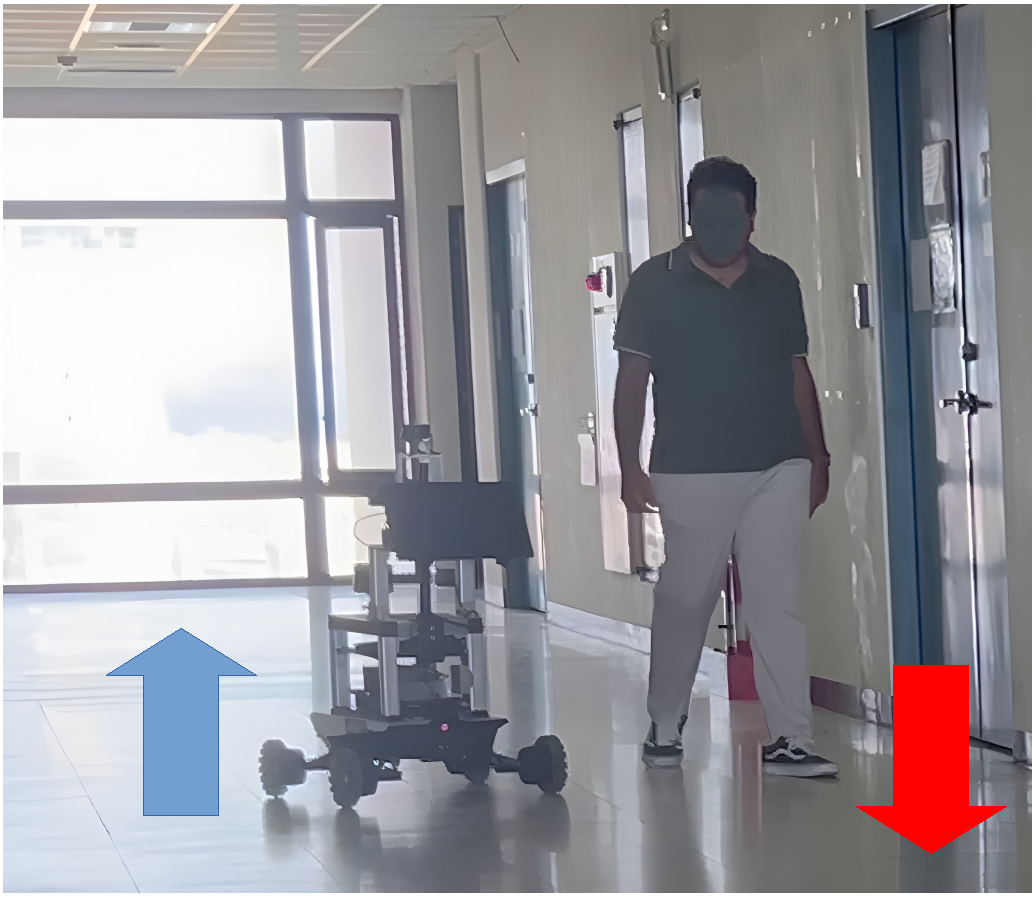}}
    \caption{
    (a) An Agilex Scout Mini is the mobile base. 
    A Zedx depth camera detects the pedestrians. 
    A 3D LiDAR measures the lateral distance. 
    \textcolor{black}{(b) The robot approaches the pedestrian head-on~\cite{Francis2025} and passes him in a trial.}
    The supplementary video shows a sample experimental run.}
    \label{fig:ExpSetup}
\end{figure}
Fig~\ref{fig:ExpSetup}(a) shows the mobile robot used during the trials.
Setting the lateral velocity to zero converts the mobile robot with Mecanum wheels to a nonholonomic differential-drive robot.
A depth camera and a 3D LiDAR detect the pedestrian and measure the relative distance and velocity vectors.
A 2D LiDAR triggers emergency shutdowns to prevent collisions.
\par
StereoLabs Zedx camera captures the pedestrian using the corresponding ROS2 package and publishes the relative position and velocity topics at a rate of 15 Hz; We use the ROS2 Humble middleware.
The 3D LiDAR identifies the pedestrian in the camera's blind spots with a rate of 10 Hz.
The Nvidia Jetson AGX Orin Developer Kit serves as the processing unit, executing ROS2 packages.
It records the robot's heading, its speed, the relative position, and the relative velocity.
\subsection{\textcolor{black}{Experimental Procedure}}\label{sec:exp:DOE}
This section describes the design of the experiments, the participants' biases, and their age and gender distributions.
Twenty-two males and ten females volunteered following announcements posted on social media groups; previous research does not support a significant gender-dependent attitude toward mobile robots~\cite{Song2025}.
Their age ranges from 23 to 41 years old, with a mean and standard deviation of 25.7 and 3.58 years, respectively.
\par
The pedestrian and the robot move toward each other in a 3.2-meter-wide hallway.
Instructions recommend that the participants walk casually and comfortably.
They report their comfort level on a 1 to 5 Likert scale using questionnaires during the trials; 1 is the most dangerous interaction and 5 stands for the most comfortable one.
\par
Because the participants are volunteers, positive bias in reporting comfort levels is expected.
The questionnaire includes additional Likert scale questions regarding the participants' trust in the safe operation of mobile robots.
We collect the volunteers' responses to these questions before the trials.
Twenty of thirty-two participants are familiar with mobile robots.
The result, with an average of 3.84 and a standard deviation of 0.93, indicates a positive bias compared to neutral populations~\cite{Naneva2020}.
\par
During the trials, an operator with a remote control navigates the robots in the hallway.
The operator moves the robot toward the end of the hall, passes the pedestrian, and returns to the hall's center.
The robot speed is constant at 1.4 m/s and 2.8 m/s, for ``R14" and ``R28" trial types, respectively.
\textcolor{black}{At lower speeds, the randomness in situation interpretation may dominate the reported discomforts and obscure the correlations.
At higher relative speeds, pedestrians' reported comfort is less individualized~\cite {Jafari2024-2-natcom}.
}
The participants believe that the robot moves autonomously and are unaware of the human operator.
\par
Each participant attends five trials that are either all R14 or all R28, resulting in $5\times16=80$ trials for each speed.
We discard 2 of 160 trials because of complete data collection failures.
Another 13 trials are dismissed only in the related analysis due to partial failure in lateral distance measurements by LiDAR.
\par
\textcolor{black}{The pedestrians may become more comfortable with the setup over consecutive trials at the same speed.
Table~\ref{tab:learning} presents comfort ratings across five consecutive trials averaged over the volunteers.
Linear regression shows R14 and R28 have a slight negative slope (-0.10, p = 0.18) 
and a slight positive slope (+0.10, p = 0.26), respectively.
The trends do not reach statistical significance.
Additionally, the aggregated regression line is flat, and indicates that pedestrians do not report higher comfort ratings in subsequent trials. 
Moreover, the relatively stable standard deviation (R14: 17-28\% and R28: 29-36\%) suggests consistent individual differences in comfort perception throughout the experiment.
Overall, there is no evidence of volunteers adapting to the test setup.
}
\par
\begin{table}[!t]
\centering
\small 
\setlength{\tabcolsep}{3pt} 
\renewcommand{\arraystretch}{1.2} 
\caption{\textcolor{black}{Comfort ratings across five consecutive trials averaged over the volunteers. Cells report the corresponding mean (absolute SD/mean$\times$100) for R14, R28, and the total.}}
\begin{tabular}{|
                  >{\centering\arraybackslash}m{2.5em}|
                  >{\centering\arraybackslash}m{2.5em}|
                  >{\centering\arraybackslash}m{2.5em}|
                  >{\centering\arraybackslash}m{2.5em}|
                  >{\centering\arraybackslash}m{2.5em}|
                  >{\centering\arraybackslash}m{2.5em}|
                  >{\centering\arraybackslash}m{2.5em}|
                  >{\centering\arraybackslash}m{2.5em}|
                 }
\hline
\rotatebox{90}{\textbf{Group}}
  & \rotatebox{90}{\textbf{Trial 1}}
  & \rotatebox{90}{\textbf{Trial 2}}
  & \rotatebox{90}{\textbf{Trial 3}}
  & \rotatebox{90}{\textbf{Trial 4}}
  & \rotatebox{90}{\textbf{Trial 5}}
  & \rotatebox{90}{\textbf{Slope}}
  & \rotatebox{90}{\textbf{p--value~}} \\
\hline\hline
R14 & 4.25 (20\%) & 4.12 (17\%) & 4.06 (28\%) & 4.00 (26\%) & 3.81 (26\%) & -0.10 &  0.18 \\
\hline
R28 & 3.44 (30\%) & 3.38 (34\%) & 3.44 (32\%) & 3
62 (36\%)& 3.81 (29\%) & +0.10 & 0.26 \\
\hline
Total & 3.84 (27\%) & 3.75 (27\%) & 3.75 (30\%) & 3.81 (31\%) & 3.81 (27\%) & 0.0 & 1.0 \\
\hline
\end{tabular}
\label{tab:learning}
\end{table}
\section{Results and Discussions}\label{sec:res}
This section presents the experimental results using boxplots, discusses the statistical significance of the correlations, and compares the variables.
Then, it designs empirical comfort estimators using the collected data, interprets their statistics, and compares their performance.
A limitation section discusses the drawbacks, challenges, and study constraints.
\par
\subsection{Subjective comfort and kinematic variables}\label{sec:res:var}
\begin{figure*}[t]
\centering
\subfloat[]{\includegraphics[width=0.32\textwidth]{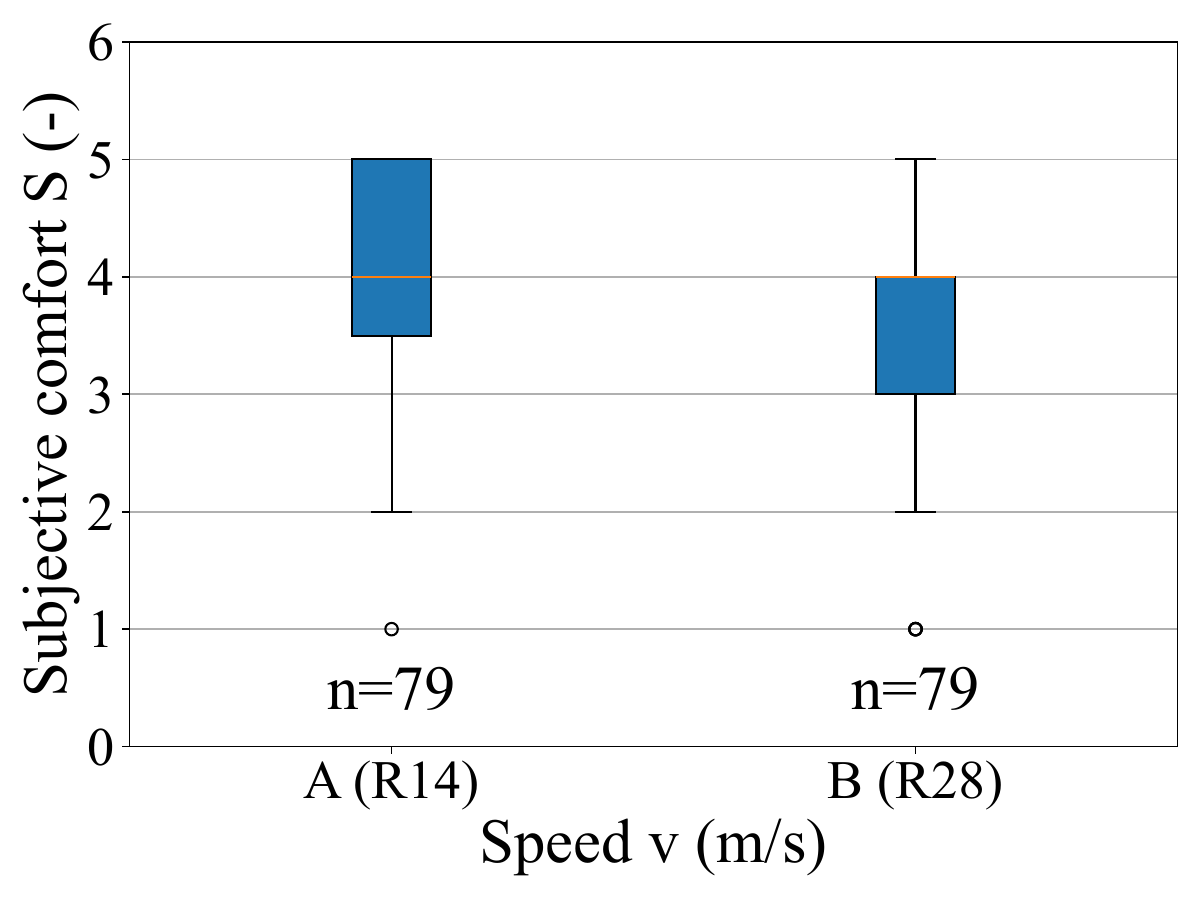}}
\subfloat[]{\includegraphics[width=0.32\textwidth]{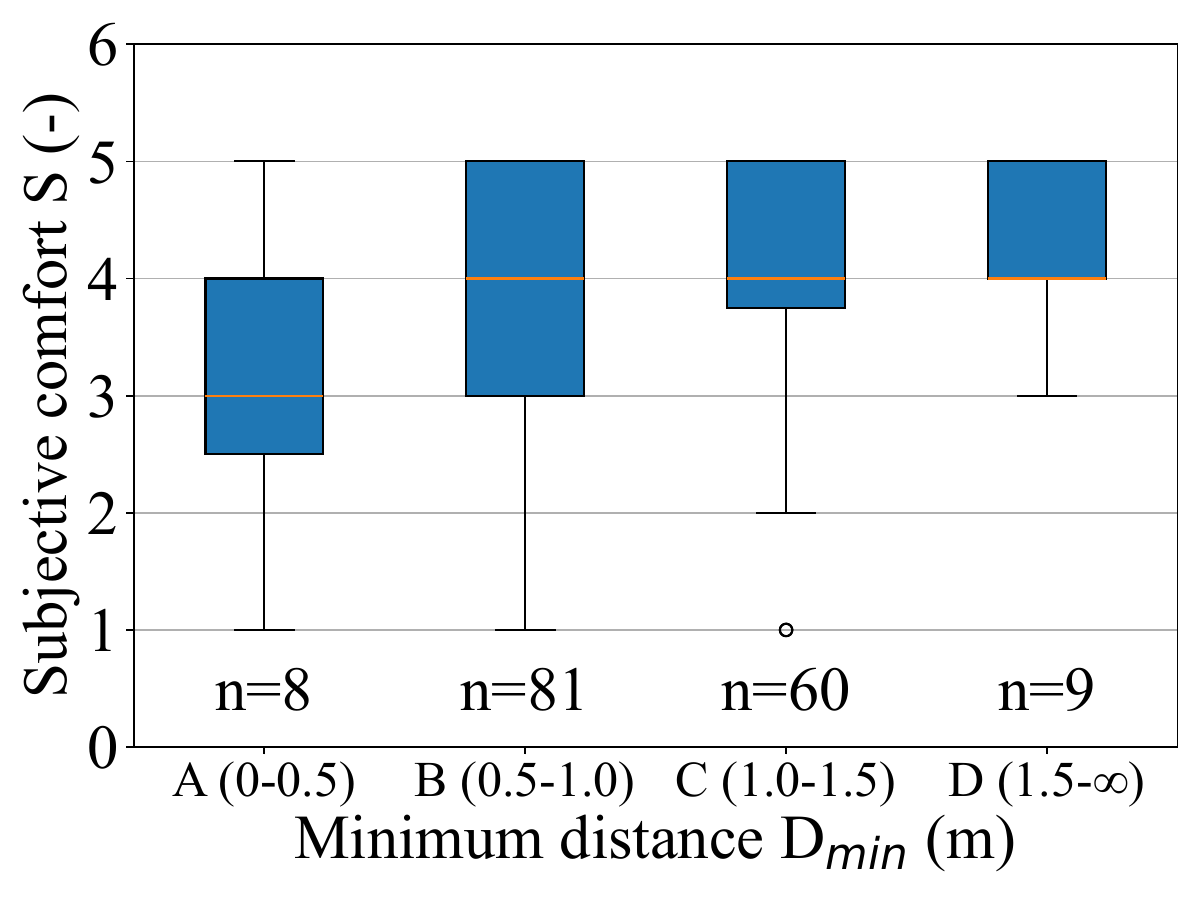}}
\subfloat[]{\includegraphics[width=0.32\textwidth]{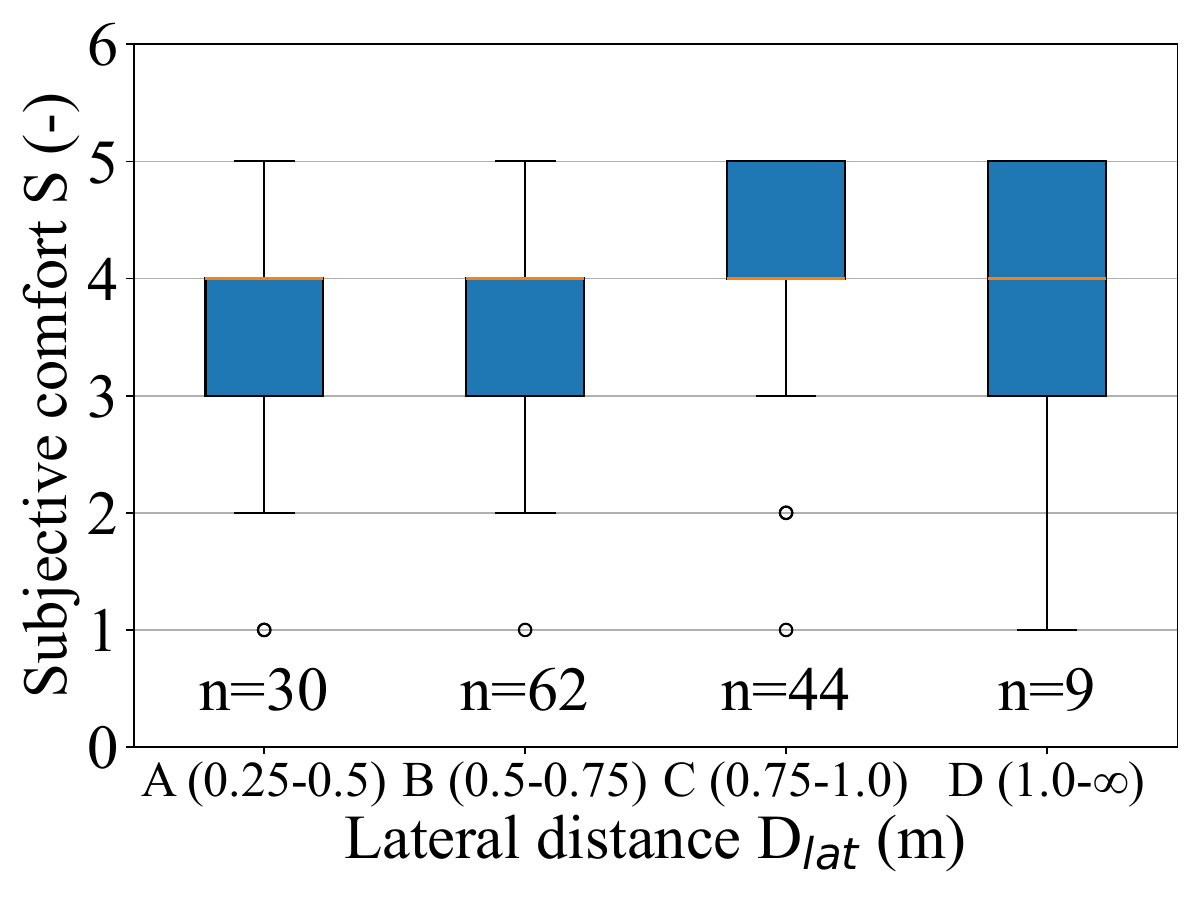}}
\\[2mm] 
\subfloat[]{\includegraphics[width=0.32\textwidth]{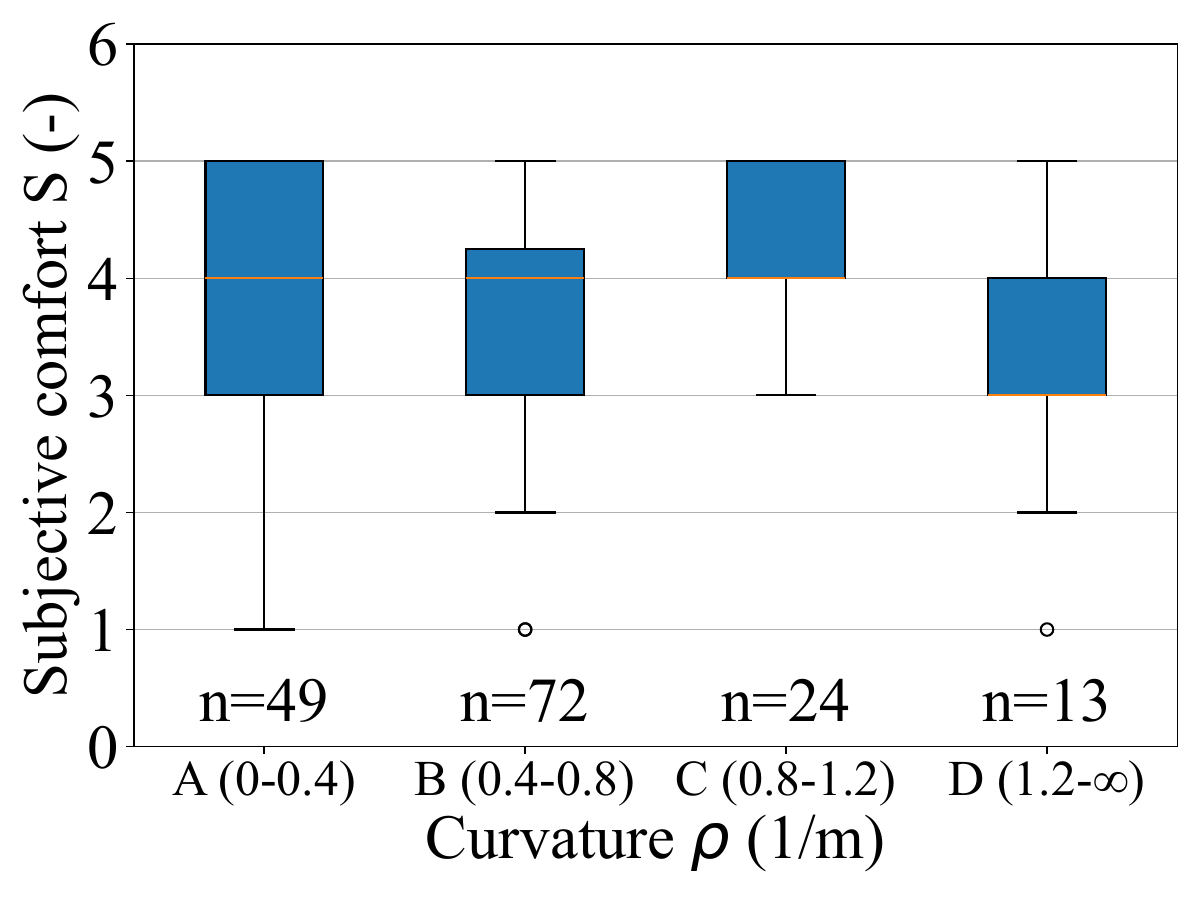}}
\subfloat[]{\includegraphics[width=0.32\textwidth]{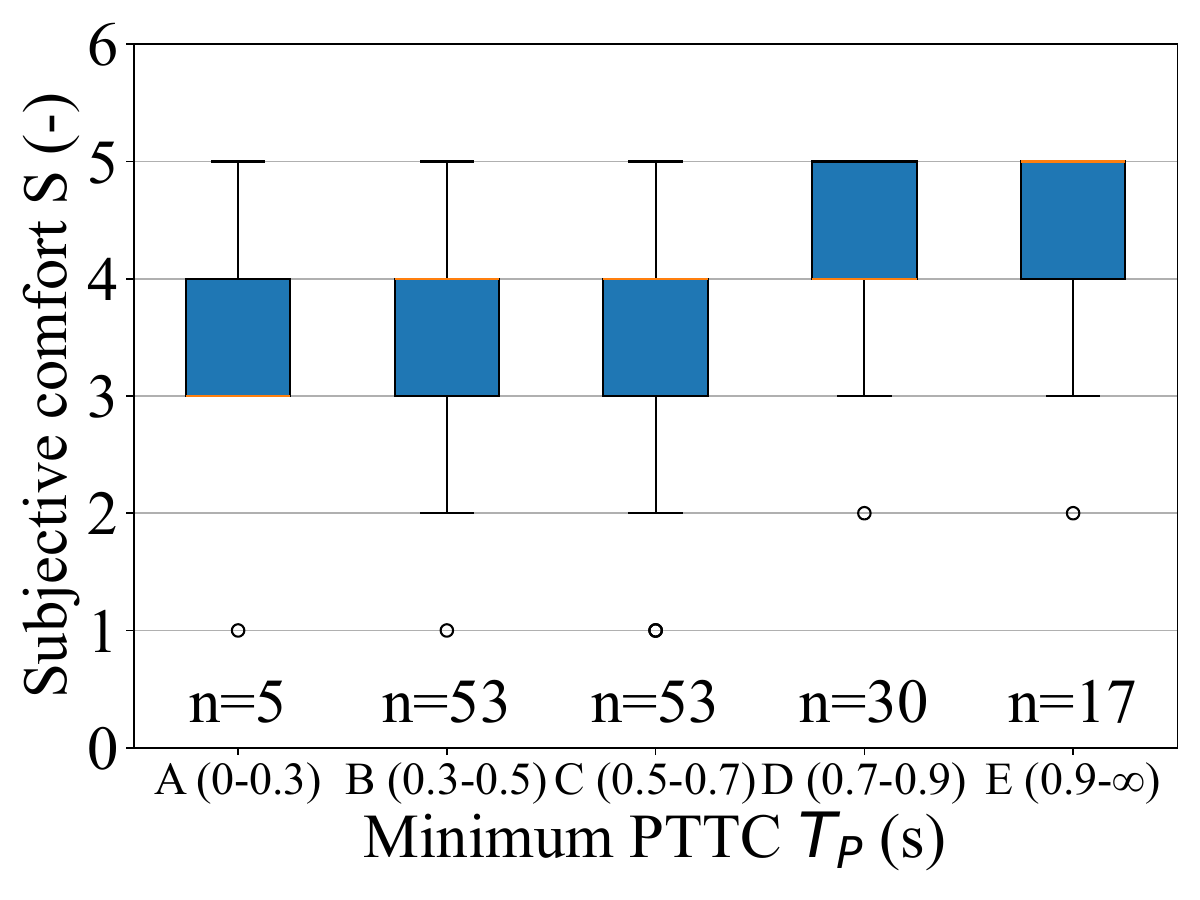}}
\subfloat[]{\includegraphics[width=0.32\textwidth]{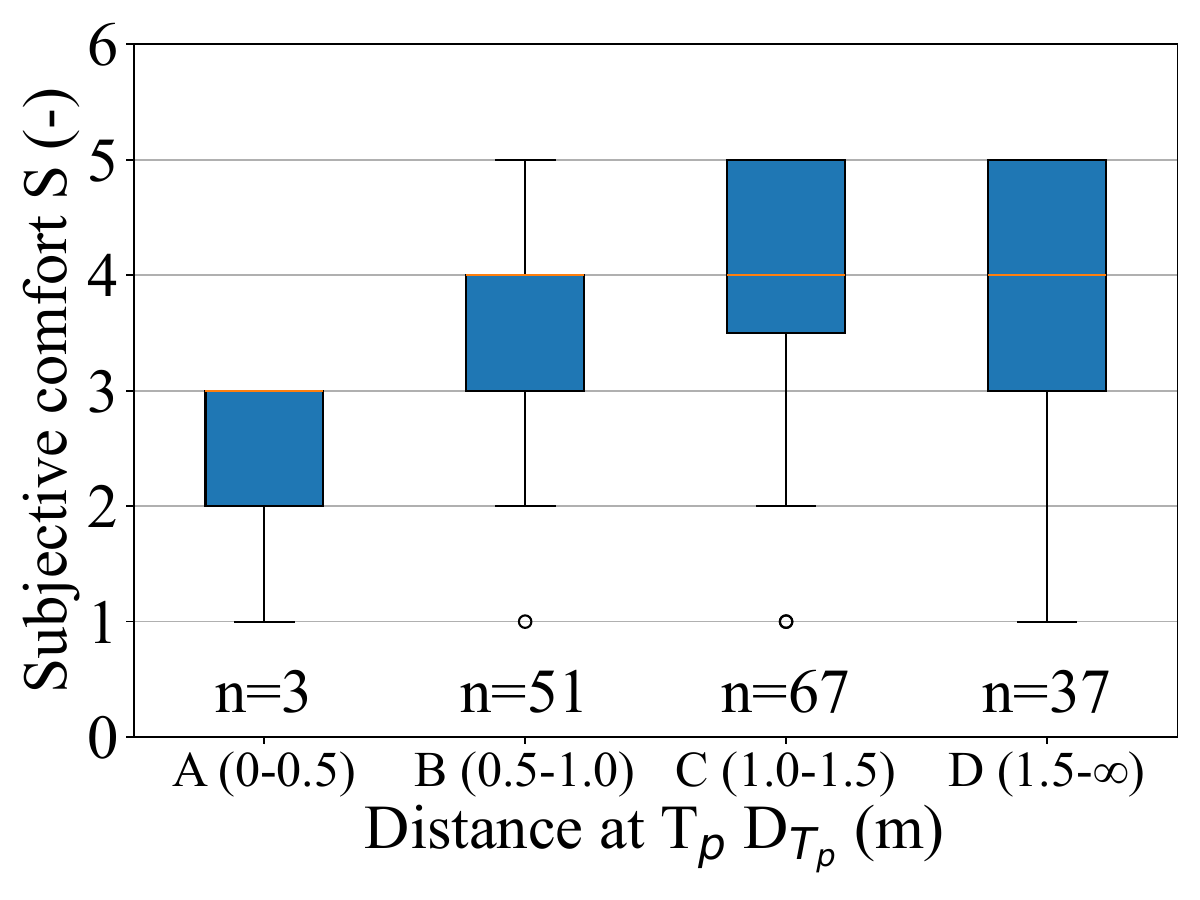}}
\caption{Visual representation of the collected data and the kinematic variable trends after binning. The y--axis is the subjective comfort $S$. The x--axis is
(a) The robot speed $v$.
(b) The minimum distance $D_{min}$. 
(c) The lateral distance $D_{lat}$.
(d) The maximum curvature $\rho$.
(e) The minimum PTTC $T_p$
(f) The distance at minimum PTTC $D_{T_P}$.}
\label{fig:boxplot_result}
\end{figure*}
This section provides insights into how pedestrians react to mobile robot motion parameters.
We visually present the experimental results through box plots to highlight and discuss the trends in the collected comfort data across the kinematic variables.
In addition, we study the comfort correlation with the variables statistically and summarize the final results in Table~\ref{tab:dcor_p}.

\par
In the following figures, the number of samples in each box is marked on the figure. 
The outliers are data points outside 1.5 times the IQR above the third quartile or below the first.
However, they are not removed from the statistical analysis.
Additionally, we employ distance correlation ($dCor$) with permutation testing (1000 iterations) for statistical analysis because it doesn't require any normality assumptions and detects nonlinear nonmonotonic relationships.
Moreover, it handles ordinal variables like Likert-scale comfort ratings where the equal-interval assumption between response categories is not necessarily correct.
\par
Fig.~\ref{fig:boxplot_result}(a) presents the reported comfort for the two speed groups, R14 and R28.
The results show that the subjective comfort $S$ decreases with robot speed $v$, confirming the study by Neggers et al.~\cite{Neggers2022-Fron}.
The comfort means (standard deviation) are 4.04 (0.94) and 3.56 (1.12) for the R14 and R28 groups, respectively.
In addition, the distance correlation categorizes the relationship between comfort and speed as moderate but statistically significant with $dCor = 0.214$ and permutation $p<0.001$.
The moderate relationship suggests speed is a contributing factor to pedestrian comfort.
\par
Fig.~\ref{fig:boxplot_result}(b) depicts the subjective comfort $S$ trend across four minimum distance bins, A to D. 
The results show that the pedestrians generally feel more comfortable when the minimum distance is higher.
Similar to speed, the statistical test confirms a moderate but significant relationship between comfort and the minimum distance with $dCor=0.219$ and permutation $p<0.011$.
The trend involves a sharp increase at low distances around 0.5 m, which flattens as the distance becomes larger.
The 0.5 m threshold suggests that robot navigation algorithms must maintain distances above 0.5 m for pedestrian comfort considerations.
\par
Fig.~\ref{fig:boxplot_result}(c) shows comfort versus lateral distance $D_{lat}$ grouped in four bins.
A nonmonotonic inverted U trend between the lateral distance and pedestrian comfort, peaking in bin C, suggests an optimal $D_{lat}$ around 0.75--1.0 m.
The trend aligns with Pacchierotti et al.'s study~\cite{Pacchierotti2006}. 
They test lateral distances of 0.2, 0.3, and 0.4 m and report that, for general users, the highest comfort ratings happen at 0.4 m.
On the contrary, the result contradicts Neggers et al.'s study reporting a monotonic increase in comfort with lateral distance~\cite{Neggers2022-Fron}, possibly due to the small sample size of Fig.~\ref{fig:boxplot_result}(c)'s last bin.
\par
Moreover, the distance correlation analysis indicates a weak and insignificant correlation, $dCor=0.167$ with a permutation $p=0.132$, contrasting with the significant relation reported by~\cite{Neggers2022-Fron}.
Perhaps the reason lies in the more strictly controlled nature of their experiments compared to our more realistic trials.
Overall, there is not enough evidence for a statistically significant relationship between subjective comfort and lateral distance.
\par
Fig.~\ref{fig:boxplot_result}(d) is the subjective comfort across four bins containing maximum curvature $\rho$ of the trajectory recorded in each trial.
The plot indicates a nonmonotonic, inverted U relationship where the comfort peaks in bin C, $0.8<\rho<1.2$, and drops in extreme curvature.
Similar to Greenberg et al.~\cite{Greenberg2025-2}, we did not find sufficient evidence of a statistically significant relation ($dCor=0.107$, $p=0.732$). 
The results suggest that maximum curvature is not a suitable sole predictor of subjective comfort.
However, Fig.~\ref{fig:boxplot_result}(d) shows patterns similar to those reported by Greenberg et al.\cite{Greenberg2025} in a previous study. 
The figure shares the inverted U shape and the existence of an optimal maximum curvature with the study.
Briefly, despite the absence of strong statistical significance, visual patterns hint at a subtle relation between the curvature and the subjective comfort.
\par
In Fig.~\ref{fig:boxplot_result}(e), the mean comfort levels progressively increase across the bins of $T_p$.
Therefore, the longer $T_p$ relates to higher comfort.
The effect, while moderate with $dCor=0.269$, is statistically significant with $p=0.002$.
The findings align with previous research using e-scooters~\cite{Jafari2024-2-natcom} demonstrating pedestrians feel more comfortable if the $T_p$ is higher.
The statistical analysis suggests that $T_p$, although not exclusive, has the strongest correlation with comfort among the studied kinematic variables.
\par
Fig.~\ref{fig:boxplot_result}(f) displays the subjective comfort trend versus the distance at the minimum PTTC, $T_p$.
The plot shows that the comfort and $D_{T_p}$ have a nonmonotonic inverted U shape relation, like the lateral distance. 
The highest comfort is apparent in bin C when $1.0\leq D_{T_p}<1.5$.
The association is modest, $dCor=0.196$, but statistically significant with permutation $p=0.033$.
Therefore, $D_{T_p}$, besides the other kinematic variables, affects pedestrian comfort.
\par
Table~\ref{tab:dcor_p} summarizes the statistical tests in this section.
Overall, the statistical and visual analysis demonstrates that pedestrian comfort is most strongly associated with the minimum PTTC $T_p$. 
The next kinematic associated variables are the speed $v$, the distance at $T_p$, and the minimum distance $D_{min}$. 
Lateral distance and maximum curvature have the weakest correlations, although they suggest visible trends.
Yet, despite the existence of better indicators, the most used pedestrian comfort metric is the minimum distance, perhaps due to the measurement's practicalities.
\begin{table}
\renewcommand{\arraystretch}{1.5}
\caption{Statistical comparison of the subjective comfort variation with the kinematic variables using distance correlation $dCor$ and corresponding permutation p--value; the permutation iteration is 1000.}
\begin{center}
\begin{tabular}{|c|c|c|c|}
\hline
\textbf{Variable} & $\boldsymbol{dCor}$ & \textbf{Permutation p} & \textbf{Sample Size} \\ 
\hline\hline
Robot speed ($v$) & 0.214 & $<$0.001 &  158 \\
\hline
Minimum distance ($D_{min}$) & 0.219 & $<$0.011 & 158 \\
\hline
Lateral distance ($D_{lat}$) & 0.167 & 0.132 & 145$^{\mathrm{a}}$ \\
\hline
Max.\ curvature ($\rho$) & 0.107 & 0.732 & 158 \\
\hline
Minimum PTTC ($T_p$) & 0.269 & 0.002 & 158 \\
\hline
Distance at min.\ $T_p$ ($D_{T_p}$) & 0.196 & 0.033 & 158 \\
\hline
\multicolumn{4}{l}{$^{\mathrm{a}}$LiDAR data were lost in some trials despite camera recordings.} \\
\end{tabular}
\label{tab:dcor_p}
\end{center}
\end{table}
\subsection{Predictors}\label{sec:res:pre}
In this section, we predict human comfort using the kinematic variables and their empirical relations with comfort visible in Fig.~\ref{fig:boxplot_result}.
The minimum distance estimator $\hat{S}_d$ only uses $D_{min}$; the minimum PTTC estimator $\hat{S}_t$ only relies on $T_p$; the composite estimator $\hat{S}_E$ combines all six kinematic variables.
The reasoning behind the estimator’s design is as follows:
\begin{itemize}
    \item The minimum distance, as it is the most widely used variable in the literature and provides a comparison baseline.
    \item The minimum PTTC, because it exhibits the strongest correlation with comfort.
    \item All variables, since each captures a distinct characteristic of human comfort, even though they are not independent and may show high collinearity.
\end{itemize}
\par
To simplify the estimations, we assume that the pedestrian is comfortable if the reported comfort level is 4 or 5 and uncomfortable otherwise.
Thus, the binary comfort index is
\begin{align}
S_E =
\begin{cases}
1, & \text{if } S \ge 4,\\
0, & \text{otherwise}
\end{cases}.
\end{align}
The estimators must predict $S_E$.
The chi-square test evaluates the predictors' reliability in detecting human comfort.
The higher the $\chi^2$ value, the better the predictions match the collected data.
Moreover, we use accuracy, precision, recall (sensitivity), specificity, and F1 (the harmonic mean of precision and recall) metrics to compare the three estimators.
The definitions are
\begin{align}
    &\text{Accuracy}=\frac{\text{TP+TN}}{\text{Total}}, \\
    &\text{Precision}=\frac{\text{TP}}{\text{TP+FP}}, \\
    &\text{Recall}=\frac{\text{TP}}{\text{TP+FN}}, \\
    &\text{Specificity}=\frac{\text{TN}}{\text{TN+FP}}, \\
    &F_1=2\times \frac{\text{Precision} \times \text{Recall}}{\text{Precision}+\text{Recall}},
\end{align}
where TP, TN, FP, and FN are the numbers of true positives, true negatives, false positives, and false negatives, respectively.
\par
\textbf{Minimum distance-based predictor:} This predictor uses Fig.~\ref{fig:boxplot_result}(b) by setting
\begin{align}
\hat{S}_d =
\begin{cases}
1, & \text{if } D_{min} \ge 1.0,\\
0, & \text{otherwise}
\end{cases}.
\end{align}
For $\hat{S}_d$, the Chi-square test result is $\chi^2=3.92$ with $p=0.0478$.
The p--value is just below the typical 0.05 threshold for statistical significance, showing that the correlation between the predictor and the binary comfort is moderate to strong.
\par
Table~\ref{tab:contingency_ShatD_SE} is the contingency table. 
Considering the odds ratio in the table, the odds of actually having $S_E=1$ are $2.28$ fold higher when $\hat{S}_d=1$ compared to when $\hat{S}_d=0$.
In other words, the odds of having a comfortable pedestrian are almost twice as high when the estimator identifies it as comfortable.
\begin{table}
\renewcommand{\arraystretch}{1.5}
\caption{Contingency table of \(\hat{S}_d\) vs \(S_E\)}
\begin{center}
\begin{tabular}{|c|c|c|}
\hline
\textbf{\( \boldsymbol{\hat{S}_d \backslash S_E}\)} & \textbf{0} & \textbf{1} \\
\hline\hline
\textbf{0} & 35 & 48 \\
\hline
\textbf{1} & 15 & 47 \\
\hline
\end{tabular}
\end{center}
\label{tab:contingency_ShatD_SE}
\end{table}
\par
\textbf{Minimum PTTC predictor:} This predictor utilizes the Fig.~\ref{fig:boxplot_result}(e) and sets
\begin{align}
\hat{S}_t =
\begin{cases}
1, & \text{if } T_{p} \ge 0.7,\\
0, & \text{otherwise}
\end{cases}.
\end{align}
Chi-square test result shows moderate association with $\chi^2=2.46$ and $p=0.12$.
The correlation between the estimator and the binary comfort is moderate. 
The tests in Section~\ref{sec:res:var} show that the correlation between the $S$ and $T_P$ is stronger than that between $S$ and $D_{min}$.
However, when converted to binary values using binning in Fig.~\ref{fig:boxplot_result}, the $T_p$ association drops significantly, possibly due to imperfect manual selection.
\par
Table~\ref{tab:contingency_Shatt_SE} presents the counts of TP, TN, FP, and FN for $\hat{S}_t$.
Based on the odds ratio, the likelihood of encountering a comfortable pedestrian $S_E=1$ is approximately 2.03 times higher when the estimator predicts $\hat{S}_t=1$, compared to when it predicts otherwise.
Put simply, when the estimator flags a case as comfortable, the odds of it truly being comfortable nearly double.
\par
\begin{table}
\renewcommand{\arraystretch}{1.5}
\caption{Contingency table of \(\hat{S}_t\) vs \(S_E\)}
\begin{center}
\begin{tabular}{|c|c|c|}
\hline
\textbf{\(\boldsymbol{\hat{S}_t \backslash S_E}\)} & \textbf{0} & \textbf{1} \\
\hline\hline
\textbf{0} & 40 & 63 \\
\hline
\textbf{1} & 10 & 32 \\
\hline
\end{tabular}
\end{center}
\label{tab:contingency_Shatt_SE}
\end{table}
\textbf{Composite predictor:} This predictor utilizes all the kinematic variables in the Fig.~\ref{fig:boxplot_result}.
However, since there are multiple variables, it requires a middle step compared to the previous predictors.
We define weights for each bin in each of the boxes of the collected data. 
The middle variable $E$ combines these weights as
\begin{align}
E=w_v+w_d+w_l+w_r+w_t+w_p,
\end{align}
where Table~\ref{tab:weights} shows the assigned weights to each bin corresponding to the kinematic variables.
\begin{table}
\renewcommand{\arraystretch}{1.5}
\caption{The assigned weights to each bin in Fig.~\ref{fig:boxplot_result}. The composite estimator combines all the variables across their internal bins.}
\begin{center}
\begin{tabular}{|c|c|c|c|c|c|c|c|}
\hline
\multicolumn{2}{|c|}{\textbf{Variable}} & {$\boldsymbol{v}$} & {$\boldsymbol{D_{min}}$} & {$\boldsymbol{D_{lat}}$} & {$\boldsymbol{\rho}$} & {$\boldsymbol{T_p}$} & {$\boldsymbol{D_{T_p}}$} \\
\hline
\multicolumn{2}{|c|}{\textbf{Weight}} & {$\boldsymbol{w_v}$} & {$\boldsymbol{w_d}$} & {$\boldsymbol{w_l}$} & {$\boldsymbol{w_r}$} & {$\boldsymbol{w_t}$} & {$\boldsymbol{w_p}$} \\
\hline\hline
\multirow{5}{*}{\textbf{Bins}} & \textbf{A} & 2 & 0 & 0 & 1 & 0 & 0 \\
\cline{2-8}
& \textbf{B} & 0 & 1 & 0 & 0 & 0 & 1 \\
\cline{2-8}
& \textbf{C} & -- & 2 & 2 & 2 & 0 & 2 \\
\cline{2-8}
& \textbf{D} & -- & 2 & 1 & 0 & 2 & 2 \\
\cline{2-8}
& \textbf{E} & -- & -- & -- & -- & 2 & -- \\
\hline
\end{tabular}
\label{tab:weights}
\end{center}
\end{table}
\par
The weights are assigned using visual inspection of how boxes present the subjective comfort.
When a box contains many data points corresponding to the highest comfort levels, the maximum score is assigned to it.
If it corresponds to mostly low comfort levels, its weight is set to zero.
If it is unclear, the weight assumes a neutral value.
\par
\begin{figure}
  \centering
  \includegraphics[width=0.48\textwidth]{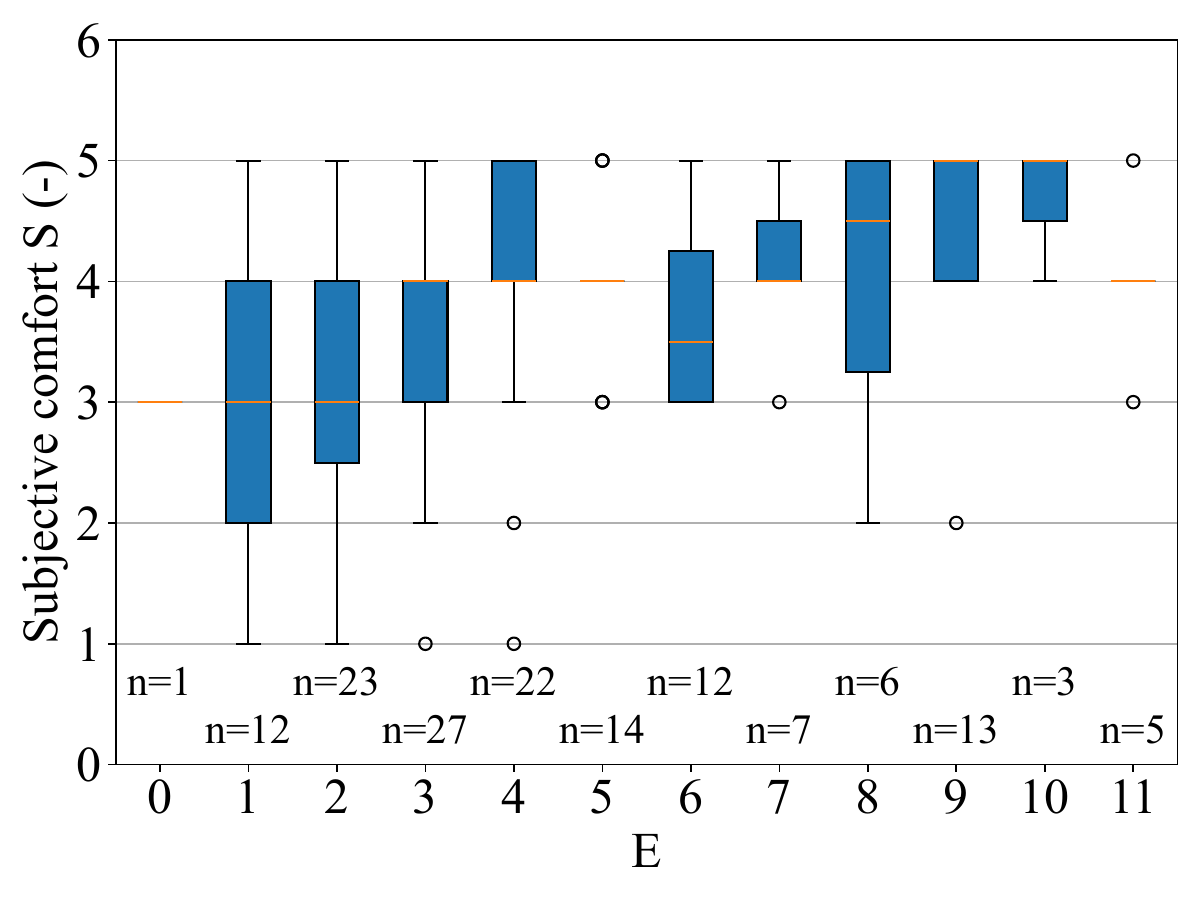}
  \caption{Comfort scores trend with the middle variable E.}
  \label{fig:boxplot_predictors}
\end{figure}

Fig.~\ref{fig:boxplot_predictors} shows $E$ trend with the reported comfort.
An almost monotonic trend relates $S$ and $E$. 
Based on the visual inspection of the boxes, we set
\begin{align}
\hat{S}_E =
\begin{cases}
1, & \text{if } E \ge 4,\\
0, & \text{otherwise}
\end{cases},
\end{align}
because most of the data points with $E<4$ have associated comfort levels below 4.
For the composite estimator $\hat{S}_E$, the Chi-square test indicates a high statistical significance compared to previous metrics with $\chi^2=11.87$ and $p=0.0006$, way below the typical threshold.
The stronger correlation indicates that the combined metric outperforms the individual variable metrics.
Therefore, despite collinearities, it is most likely true that each variable captures a distinct aspect of human comfort.
\par
Table~\ref{tab:contingency_ShatE_SE} is the contingency table for the estimator $\hat{S}_E$.
The odds ratios of the $\hat{S}_E$ are 3.67.
For the composite predictor, it means that when the algorithm flags comfort, $\hat{S}_E=1$, the odds that the pedestrian is actually comfortable, $S_E=1$, are 3.67 times the odds when it does not flag comfort.
For the other methods, the odds are significantly lower, showing the advantage of the combined method over the others, especially in reward/punishment trajectory evaluating algorithms.
\begin{table}
\renewcommand{\arraystretch}{1.5}
\caption{Contingency table of \(\hat{S}_E\) vs \(S_E\)}
\begin{center}
\begin{tabular}{|c|c|c|}
\hline
\textbf{\(\boldsymbol{\hat{S}_E \backslash S_E}\)} & \textbf{0} & \textbf{1} \\
\hline\hline
\textbf{0} & 32 & 31 \\
\hline
\textbf{1} & 18 & 64 \\
\hline
\end{tabular}
\end{center}
\label{tab:contingency_ShatE_SE}
\end{table}
Table~\ref{tab:estimator_comparison} compares the estimator's performance.
It demonstrates that $\hat{S}_E$, the composite index, outperforms the other estimators in four of the five metrics: the accuracy, the precision, the sensitivity (recall), and, as a result $F_1$ score.
However, in specificity, the other two estimators perform better.
\par
While all three models show relatively high recall (75-78\%), indicating they successfully identify most positive cases, they have moderate specificity at best, particularly $\hat{S}_d$ and $\hat{S}_t$, struggling in correctly identifying negative cases.
Table~\ref{tab:estimator_comparison} suggests that $\hat{S}_E$ has the best balance between detecting true positives and avoiding false alarms according to the $F_1$ score.
\begin{table}
\renewcommand{\arraystretch}{1.2}
\setlength{\tabcolsep}{3.5 pt}
\caption{Estimator comparison using performance metrics for classification.}
\begin{center}
\small
\begin{tabular}{|c|c|c|c|c|c|}
\hline
\textbf{Estimator} & \textbf{Accuracy} & \textbf{Precision} & \textbf{Recall} & \textbf{Specificity} & $\boldsymbol{F}_\textbf{1}$\\
\hline\hline
\textbf{$\hat{S}_E$} & 0.662            & 0.674              & 0.781            & 0.508             & 0.723 \\
\hline
\textbf{$\hat{S}_d$} & 0.566            & 0.495             & 0.758             & 0.422             & 0.599 \\
\hline
\textbf{$\hat{S}_t$} & 0.497            & 0.337             & 0.762              & 0.388            & 0.467 \\
\hline
\end{tabular}
\label{tab:estimator_comparison}
\end{center}
\end{table}
\subsection{Limitations}\label{sec:res:lim}
Despite providing insights into robot-pedestrian interaction kinematics and designing potentially useful comfort estimators, the study faces a few methodological constraints.
Although the participants' group size and bias are comparable with those of similar research, a comprehensive study requires a larger group of participants and a more gender and age-balanced population.
In addition, the volunteer sampling leads to positive attitudes toward mobile robots, distorting the comfort responses distributions.
Moreover, the reliance on post-trial measures (questionnaires) adds additional noise to the reported comfort due to the randomness in participants' judgments.
\textcolor{black}{Although the pedestrians did not adapt to the test setup, a randomized trial order better avoids such concerns.}
\par
\textcolor{black}{In addition, the same dataset used in analysis evaluates the predictors.
Since even a human hand crafted detector is a learning algorithm, a rigorous evaluation must be done using a train-test split.}
\par
Employing only two speeds is another limitation. 
Multiple intermediate speed steps yield a more robust correlation analysis in future works.
Furthermore, the remote-controlled robot exhibits patterns that do not match an autonomous mobile robot.
Thus, the distribution of the studied kinematic variables may differ in autonomous settings.
In addition, the manual binning and threshold setting obscure the underlying patterns and damage the prediction power.
For instance, the predictor based on $\hat{S}_t$ underperforms relative to $\hat{S}_d$ despite $T_p$'s higher correlation with comfort compared to $D_{min}$.
\section{Conclusions}\label{sec:con}
The paper quantifies pedestrian comfort in human-robot interactions in hallway trials.
Six kinematic variables are statistically evaluated, including the popular minimum distance to and the recent PTTC.
Experiments rank their correlation with pedestrian comfort.
We empirically design a composite comfort predictor using all the introduced kinematic variables and compare its performance with single variable predictors.
\par
Our study shows that the minimum PTTC has the highest correlation with walkers' comfort among the studied variables.
The minimum distance and the robot speed are the next immediate followers.
Regarding the designed predictors, the composite predictor outperforms the single-variable predictors, i.e., the minimum distance and the PTTC estimators.
The composite estimator achieves an $F_1$ score of 0.72 and an odds ratio of 3.67.
Thus, trajectories it flags as comfortable have, by nearly fourfold, higher odds of matching human-reported comfort compared to when it doesn't flag.
Path-planning algorithms can utilize this predictor in their cost function.
Namely, the robot assigns higher rewards to paths the composite model deems comfortable and generates socially compliant movements.
\par
\textcolor{black}{Three directions for future work are identified. 
First, we extend our model to more diverse environments and participant groups to further assess robustness.
Second, we intend to apply the composite predictor to a path-planning framework and evaluate pedestrian feedback in fully autonomous scenarios.
Third, we explore machine learning approaches to derive comfort predictions from large-scale pedestrian trajectory datasets.}
\bibliographystyle{IEEEtran}
\bibliography{references}
\end{document}